\documentclass{article}

\PassOptionsToPackage{numbers, compress}{natbib}

\usepackage[main, final]{neurips_2026}

\usepackage[utf8]{inputenc} 
\usepackage[T1]{fontenc}    
\usepackage{hyperref}       
\usepackage{url}            
\usepackage{booktabs}       
\usepackage{amsfonts}       
\usepackage{nicefrac}       
\usepackage{microtype}      
\usepackage{xcolor}         

\usepackage{amsmath}
\usepackage{graphicx}
\usepackage{wrapfig}
\usepackage{caption}
\usepackage{multirow}
\usepackage{makecell}
\usepackage{diagbox}

\usepackage{amssymb}
\usepackage{amsthm}

\usepackage{enumitem}

\title{UniReFP: Robust Unified Fingerprinting for Vision Models against Cross-Task Repurposing Attacks}

\author{
    Ziye Geng$^{1}$, \quad
    Guang Yang$^{2}$, \quad
    Yihang Chen$^{1}$, \quad
    Lingfeng Yao$^{1}$, \\
    \textbf{Xing Shi}$^{3}$\textbf{,}\quad
    \textbf{Miao Pan}$^{1}$\textbf{,} \quad
    \textbf{Xin Fu}$^{1}$\textbf{,} \quad
    \textbf{Changqing Luo}$^{1}$ \\
    $^1$University of Houston \quad
    $^2$Virginia Commonwealth University \\
    $^3$University of California, Riverside \\
}

\begin{document}

\maketitle

\begin{abstract}
We identify, for the first time, a new model modification attack---the cross-task model repurposing attack---that can render fingerprints generated by existing model fingerprinting approaches inapplicable. To address this challenge, we propose {UniReFP}---a novel unified model fingerprinting framework applicable to diverse vision models across different tasks---that utilizes ownership evidence in a shared class-level semantic space rather than original task-specific output space, enabling effective ownership verification when a protected model is repurposed across tasks with heterogeneous output formats. Specifically, UniReFP constructs a surrogate classifier by repurposing the protected model, enabling fingerprints to be generated independently of the original task-dependent output space of the protected model. Moreover, it optimizes grouped fingerprints to encode ownership evidence: fingerprints within the same group are encouraged to induce consistent semantic responses and intermediate representations on the surrogate classifier, while producing dispersed responses on independently trained reference classifiers. During verification, UniReFP uses a task-aware semantic projection mechanism to map heterogeneous suspect-model outputs into unified semantic-presence vectors, and determines ownership by measuring inner-group consistency. Extensive experimental results validate the effectiveness of UniReFP, consistently showing superior performance under cross-task repurposing attacks, commonly studied model modification attacks, and even their combinations.
\end{abstract}

\section{Introduction}

Model fingerprinting has gained increasing attention for protecting the ownership of vision models that have achieved remarkable success across many vision tasks, including image classification, semantic segmentation, and image generation \citep{MFDN22,FDNN22,TMFA22,MCGF24,li2024figan, teng2025fddpm,nie2023aigm}. It leverages inherent characteristics of a protected model to generate distinctive fingerprints, and performs ownership verification by querying black-box suspect models with these fingerprints \citep{IPIP21,TMFA22,UWSD24,MCGF24,BUNNF25}.  Particularly, prior work \citep{yang26anafp,UWSD24, QURD25, shao2025fit, geng2026irisfp} has extensively investigated robust fingerprinting against commonly-studied model modification attacks, such as fine-tuning, pruning, knowledge distillation, and their combinations, which adversaries may intentionally apply to stolen protected models to evade ownership verification.

In this paper, we identify, for the first time, a new model modification attack, called the \emph{cross-task model repurposing attack}, that compromises the intellectual property (IP) of protected vision models. Modern vision models can be typically viewed as a backbone feature extractor (referred to as the \emph{backbone} throughout this paper) followed by a task-specific prediction module (referred to as the \emph{head}). In modern vision architectures, the backbone usually contains most of the model parameters and most of the learned visual knowledge, while the task-specific head is comparatively lightweight. After training on vision datasets, the backbone learns rich and transferable visual representations that can be reused across diverse downstream tasks. Instead of training separate models from scratch for every task, an adversary can thus repurpose a protected model for a target task by reusing its backbone, attaching a new task-specific head, and lightly fine-tuning the resulting model. For example, a model trained for image classification can later be repurposed for semantic segmentation or object detection simply by replacing the prediction head. Since repurposing a well-trained model from its original task to a target task can significantly improve data efficiency~\citep{zamir2018taskonomy,vasconcelos2022proper} by providing strong initialization for training the repurposed model~\citep{lin2017feature,he2017mask,li2023aligndet}, such task-specific models have become attractive targets for adversaries seeking to reuse their backbones.

Existing model fingerprinting approaches \citep{IPIP21,QURD25,UWSD24,yang26anafp}, even including prior task-agnostic fingerprinting methods \citep{MVAM22, yang26liteguard}, are primarily developed under the assumption that adversaries perform commonly-studied model modifications to alter protected models without replacing their prediction heads. Under this assumption, the protected and suspect models share the same output format, allowing ownership evidence to be directly compared within a shared task-specific output space. Hence, fingerprints generated from a protected model remain directly verifiable after commonly-studied model modifications. For example, for image classification models, verification is usually performed by checking whether the suspect model predicts the same fingerprint labels as the protected model. 

However, this assumption does not hold under cross-task repurposing attacks. In such attacks, the adversary repurposes a protected model for a different task by replacing with a new task-specific head. This modification fundamentally alters the model’s output head, thereby rendering fingerprints generated by existing approaches inapplicable. More specifically, the suspect model no longer produces outputs in the original task-specific format, but instead returns task-dependent outputs determined by the target task, such as image-level label predictions, detection instances, or dense segmentation masks. Hence, ownership evidence defined in the original output space can no longer be directly applied to verify a repurposed model with a different output format. This distinguishes cross-task model repurposing attacks fundamentally from commonly-studied model modifications.

\noindent \textbf{Main question}. Based on the discussion so far, we have the following research question: \textit{Is it possible to design a new unified model fingerprinting approach that is applicable to diverse vision tasks and remains robust to model modification attacks, including both commonly-studied modifications and cross-task repurposing?}

\vspace{-0.1in}
\subsection{Our Answer}
\label{subsec:result}
In this paper, we give an affirmative answer to the above question.
Our key insight is that cross-task model repurposing changes the task-specific output format of a protected model, while potentially preserving transferable semantic knowledge inherited from the protected model. Therefore, ownership evidence should not be tied to the original task output format, but instead be defined in a shared semantic space where heterogeneous vision outputs can be compared through their class-level semantic content. We instantiate this principle using a semantic class presence vector, which provides a task-unified ownership signal by abstracting away output-format differences while retaining the semantic knowledge associated with the protected model.

\textbf{Our Design:} We propose UniReFP, a unified fingerprinting framework for ownership protection under cross-task model repurposing attacks. Instead of relying on the original task-specific output space, UniReFP leverages ownership evidence in a shared class-level semantic space, enabling heterogeneous outputs from repurposed models across different tasks to be compared through their class-level semantic content. To achieve this, three key challenges need to be addressed. First, fingerprints must be generated in a form that is independent of the raw output space of the protected model. UniReFP addresses this by constructing a surrogate classifier defined over the shared class-level semantic space, allowing fingerprints to be optimized through a unified class-level semantic space. Second, the resulting evidence must be distinctive to the protected model lineage rather than shared by independently trained models. UniReFP therefore optimizes grouped fingerprints to induce consistent semantic responses and intermediate representations on the surrogate classifier, while encouraging dispersed responses on independently trained reference classifiers. Third, ownership must remain verifiable when suspect models return heterogeneous outputs. UniReFP uses a task-aware semantic projection mechanism to project different task outputs into unified semantic-presence vectors, and verifies ownership by checking the inner-group consistency preserved in this unified semantic space.

\section{Problem Formulation}
\label{sec:threat}

\begin{wrapfigure}{r}{0.55\textwidth}
  \vspace{-10pt}
  \centering
    \captionsetup{belowskip=1pt,aboveskip=2pt}
  \includegraphics[width=0.55\textwidth]{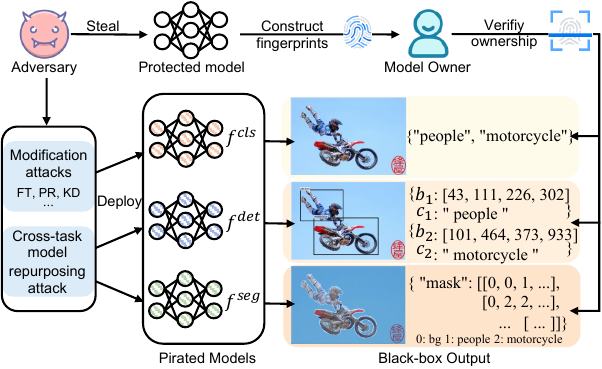}
  \caption{The considered fingerprinting scenario.}
  \label{fig:threatmodel}
  \vspace{-12pt}
\end{wrapfigure}

We consider a typical model fingerprinting scenario involving two parties: a model owner (who also acts as the verifier) and an adversary, as shown in Fig.~\ref{fig:threatmodel}. Specifically, the model owner first trains a vision model $f^{(t_s)}$, represented by $f^{(t_s)}(x)=\mathcal{H}^{(t_s)}\!\left(\mathcal{G}(x)\right)$, for a specific task $t_s$ over a dataset
$\mathcal{D}=\{(x_n, y_n^{(t_s)})\}_{n=1}^{|\mathcal{D}|}$,
where $x_n\in\mathcal{X}$ is an input image,
$y_n^{(t)}$ is the corresponding annotation for task $t_s$, $\mathcal{G}(x)$ is the backbone taking image $x$ as input, and $\mathcal{H}^{(t_s)}(\cdot)$ denotes the task-specific head for task $t_s$, then generates a set of fingerprints, which are kept confidential, to protect model ownership, and finally deploys $f^{(t_s)}$ to provide services. In contrast, the adversary first steals the protected model $f^{(t_s)}$, then applies model modification attacks to evade ownership verification, and finally deploys the resulting pirated model in a black-box setting, offering access through APIs. Upon identifying a suspect model $f_v$, the verifier (model owner) uses the generated fingerprints to query $f_v$ and compares the returned outputs with the expected responses to determine ownership.

\textbf{Cross-task model repurposing attacks.}
In addition to the model modification operations commonly studied in prior work, the adversary may launch a cross-task model repurposing attack. In this attack, the adversary constructs a new model for a target task $t_a$, where $t_a \neq t_s$, by replacing the original head $\mathcal{H}^{(t_s)}(\cdot)$ with a new task-specific head $\mathcal{H}^{(t_a)}(\cdot)$, while retaining the backbone $\mathcal{G}(\cdot)$ of the protected model. Optional modules may be inserted between the backbone and the new head to facilitate the transfer, and the resulting model is thus expressed by $f^{(t_a)}(x)=\mathcal{H}^{(t_a)}\!\left(\Psi\!\left(\mathcal{G}(x)\right)\right)$, where $\Psi(\cdot)$ represents any task-adaptation components, such as intermediate necks, feature aggregation modules, or decoders. The adversary then further fine-tunes this repurposed version $f^{(t_a)}$ to better fit the target task. For simplicity, we use $f^{(t_a)}$ to represent the final repurposed model obtained after fine-tuning.

\textbf{Design goals.}
Our goal is to design a unified model fingerprinting framework applicable to vision models across different tasks, providing ownership protection against not only commonly-studied model modification attacks but also cross-task repurposing attacks. Therefore, the designed fingerprinting approach should satisfy the following properties.

\begin{itemize}[left=0pt]
     \item \textit{Robustness to model modification attacks.} Fingerprints should remain effective for ownership verification even when an adversary applies model modification attacks—including commonly-studied operations and cross-task model repurposing—to alter protected models.

     \item \textit{Cross-task uniqueness against independently trained models.} Fingerprints should be able to effectively distinguish pirated models derived through model modification attacks—including both commonly-studied operations and cross-task model repurposing—from independently trained models, even when they share architectures, datasets, or the same class-level semantic space.

     \item \textit{Cross-task verification operations.} Fingerprints should remain verifiable across diverse task-specific vision models with heterogeneous output formats.
\end{itemize}

\section{Methodology}
\label{sec:method}
\subsection{Overview}
\label{sec:method_overview}
 \vspace{-5pt}
\begin{figure*}[htbp]
    \centering
    \includegraphics[width=0.99\linewidth]{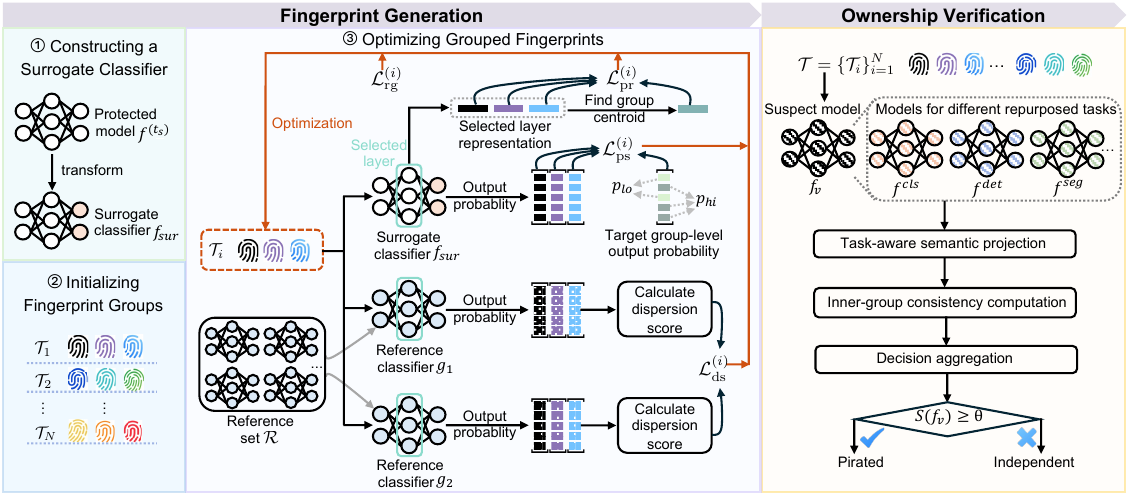}
    \caption{Overview of UniReFP.}
\label{fig:overview}
\vspace{-10pt}
\end{figure*}

We propose \textbf{UniReFP}, a unified model fingerprinting framework applicable to diverse vision models across different tasks, designed to protect model ownership against cross-task repurposing attacks. As shown in Fig.~\ref{fig:overview}, UniReFP consists of two processes: fingerprint generation and ownership verification, following the general paradigm adopted by most existing model fingerprinting approaches.

During fingerprint generation, we first repurpose a protected model into a surrogate classifier by retaining its transferable backbone and attaching a newly introduced classifier head over the underlying class-level semantic space, thereby enabling a unified fingerprint generation process for diverse protected vision models across different tasks. 
Then, we initialize fingerprint groups by randomly sampling and grouping inputs from the dataset used to train the protected model and attaching randomly initialized perturbations to each sample. 
Finally, we optimize these grouped fingerprints using the surrogate classifier and an independent reference classifier set. Particularly, the perturbations are jointly optimized to (i) enforce group consistency on the surrogate classifier, (ii) promote group dispersion on independently trained reference classifiers, and (iii) constrain perturbation magnitude via regularization. This design enables the generated fingerprints to remain effective against cross-task model repurposing attacks.

During ownership verification, we evaluate a black-box suspect vision model, which may have been derived from the protected model through cross-task model repurposing or commonly studied model modifications, using a three-step verification process.
First, for each fingerprint input, we query the suspect model to obtain its task-specific output and then project this output into a unified semantic-presence vector. Second, we compute an inner-group consistency score for each fingerprint group based on these projected semantic-presence vectors. Third, we aggregate the resulting group-level consistency scores across all fingerprint groups to obtain an overall ownership score, which is then compared with a predefined threshold to determine whether the target model is pirated or independently trained.

\subsection{Fingerprint Generation}

The fingerprint generation consists of three steps: constructing a surrogate classifier, initializing fingerprint groups, and optimizing grouped fingerprints. The details of each step are described below.

\subsubsection{Constructing a Surrogate Classifier} 
\label{method:surrogate}

Since UniReFP verifies ownership by comparing suspect-model responses in a unified class-level semantic space, the generated fingerprints should also encode distinctive and stable evidence in the same semantic space.
Thus, we introduce a surrogate classifier to enable this semantic space fingerprint generation. Recall that a protected model $f^{(t_s)}$, originally designed for task $t_s$ and trained on a dataset $\mathcal{D}$ with an underlying class-level semantic space of $C$ classes, can be expressed as
$f^{(t_s)}(x)=\mathcal{H}^{(t_s)}\!\left(\mathcal{G}(x)\right)$. To generate fingerprints for $f^{(t_s)}$, we first construct a surrogate classifier
$f_{\mathrm{sur}}(x)=\mathcal{H}_{\mathrm{sur}}\!\left(\mathcal{G}(x)\right)$ by repurposing $f^{(t_s)}$, where $\mathcal{H}_{\mathrm{sur}}(\cdot)$ is a newly introduced classifier head. This transformation enables a unified fingerprinting paradigm applicable to protected vision models across different types of tasks. It is worth noting that when the protected model is already a classifier, it is used directly for fingerprint generation without any modifications. For other vision models, we train the surrogate classifier head using image-level class-presence labels derived from the original annotations in $\mathcal{D}$. More specifically, these labels are obtained by identifying semantic classes present in each image from corresponding task-specific annotations---for example, from bounding-box category labels in object detection annotations, or from pixel-level semantic masks in semantic segmentation annotations.

\subsubsection{Initializing Fingerprint Groups}

To construct fingerprint groups, we first initialize them by randomly sampling  $Q$ inputs from $\mathcal{D}$ and dividing them into $N$ groups, each containing $M$ samples. Let
$\mathcal{S}=\{\mathcal{S}_i\}_{i=1}^{N}$, where $\mathcal{S}_i=\{x_{i1},\ldots,x_{ij},\ldots,x_{iM}\}$ is the $i$-th group with $x_{ij}$ being the $j$-th input in that group. Then, we randomly initialize trainable perturbations for each sample and construct the initial fingerprints as $\hat{x}_{ij}=x_{ij}+\delta_{ij}$, where $\delta_{ij}$ is a perturbation. This results in a grouped fingerprint set
$ \mathcal{T}=\{\mathcal{T}_i\}_{i=1}^{N}$ and a corresponding perturbations set $\Delta=\{\Delta_i\}_{i=1}^{N}$, where $\mathcal{T}_i=\{\hat{x}_{i1},\hat{x}_{i2},\ldots,\hat{x}_{iM}\}$ and $\Delta_i=\{\delta_{ij}\}_{j=1}^{M}$ represent the fingerprints and their associated perturbations for the $i$-th group, respectively.

\subsubsection{Optimizing Grouped Fingerprints}

To realize UniReFP’s design goals, we optimize the trainable perturbations to generate effective grouped fingerprints, and different fingerprint groups are optimized independently. First, to ensure that each fingerprint group exhibits stable and distinctive behavior on the protected model, we enforce group consistency at two complementary levels: semantic and representation. Semantic-level consistency encourages all fingerprints within a group to induce consistent outputs from the surrogate classifier, while representation-level consistency constrains their intermediate features to remain close in the model’s representation space. These two forms of consistency serve complementary roles: the former shapes the group-level semantic presence evidence used for verification, while the latter reinforces this semantic evidence with aligned feature representations that are more likely to persist under cross-task model repurposing attacks. Second, while group consistency aligns semantic presence within a group, it does not guarantee that independently trained models will behave differently on the same fingerprint group. To enhance separability from independently trained models, we introduce a set of reference classifiers and explicitly encourage dispersed group-level responses across them.
Finally, to maintain visual plausibility and ensure that each fingerprint remains close to its corresponding sampled input, we impose a regularization constraint on the perturbation magnitude.

Therefore, we define the loss function for a fingerprint group $\mathcal{T}_i$ by
\begin{equation}
\mathcal{L}^{(i)}
=\mathcal{L}^{(i)}_{\mathrm{ps}} + \mathcal{L}^{(i)}_{\mathrm{pr}} - \mathcal{L}^{(i)}_{\mathrm{ds}} + \lambda \mathcal{L}^{(i)}_{\mathrm{rg}},
\label{eq:overall_objective}
\end{equation}
where $\mathcal{L}^{(i)}_{\mathrm{ps}}$ enforces group consistency in the semantic output space of the surrogate classifier, $\mathcal{L}^{(i)}_{\mathrm{pr}}$ enforces group consistency in its representation space, $\mathcal{L}^{(i)}_{\mathrm{ds}}$ encourages dispersion on independently trained reference models, and $\mathcal{L}^{(i)}_{\mathrm{rg}}$ regularizes the magnitude of the perturbations. In the following, we describe each loss term in detail.

\textbf{Semantic-level group consistency $\mathcal{L}^{(i)}_{\mathrm{ps}}$.}
We propose to realize semantic-level consistency by encouraging all fingerprints within the same group to induce a shared semantic response from the surrogate classifier. Let
$p_{ij} \in (0,1)^C$
denote the output probability vector of the surrogate classifier $f_{sur}$ for the $j$-th fingerprint $\hat{x}_{ij}$ in $\mathcal{T}_i$, where $[p_{ij}]_c$ represents the predicted probability that an object belonging to the $c$-th semantic class is present in $\hat{x}_{ij}$.

For each group $\mathcal{T}_i$, we randomly select a class subset $\mathcal{C}^{\mathrm{sel}}_i \subseteq \{1, 2, \dots,C\}$, and define a target group-level output probability vector $t_i \in (0,1)^C$ by $[t_i]_c =
\begin{cases}
p_{\mathrm{hi}}, & c \in \mathcal{C}^{\mathrm{sel}}_i\\
p_{\mathrm{lo}}, & c \notin \mathcal{C}^{\mathrm{sel}}_i
\end{cases}
$, where $p_{\mathrm{hi}}$ and $p_{\mathrm{lo}}$ are predefined constants, with $0 < p_{\mathrm{lo}} < p_{\mathrm{hi}} < 1$, specifying high and low presence probabilities for selected and non-selected classes, respectively. Then, we encourage all fingerprints in $\mathcal{T}_i$ to match this target by minimizing $\mathcal{L}^{(i)}_{\mathrm{ps}}
=
\frac{1}{M}\sum_{j=1}^{M}
\mathrm{BCE}(p_{ij}, t_i)$,
where $M$ is the number of fingerprints in $\mathcal{T}_i$, and $\mathrm{BCE}(\cdot,\cdot)$ is the binary cross-entropy loss. Minimizing $\mathcal{L}^{(i)}_{\mathrm{ps}}$ aligns all fingerprints within a group to the same target probability vector, thereby inducing a consistent group-level semantic response.

\textbf{Representation-level group consistency  $\mathcal{L}^{(i)}_{\mathrm{pr}}$.}
We further enforce representation-level consistency on the surrogate classifier. In cross-task model repurposing attacks, the task-specific head may change significantly, while the transferable backbone representation is largely preserved. Therefore, encouraging fingerprints within the same group to share similar feature representations in the representation space helps stabilize their group-level behavior after repurposing.

To quantify this consistency, let $\mathcal{K}$ denote a set of selected intermediate layers within the backbone of the protected model, and $z_{ij}^{(\ell)}$ denotes the representation at layer $\ell \in \mathcal{K}$ for the $j$-th fingerprint $\hat{x}_{ij}$ in group $\mathcal{T}_i$. 
For each group $\mathcal{T}_i$, the group centroid at layer $\ell$ is defined as $\bar{z}_{i}^{(\ell)} = \frac{1}{M}\sum_{j=1}^{M} z_{ij}^{(\ell)}$.
The representation-level group consistency loss is defined by $\mathcal{L}^{(i)}_{\mathrm{pr}}
=
\frac{1}{|\mathcal{K}|M}
\sum_{\ell \in \mathcal{K}}
\sum_{j=1}^{M}
\left(1-\cos\!\left(z_{ij}^{(\ell)}, \bar{z}_{i}^{(\ell)}\right)\right)$,
where $\cos(\cdot,\cdot)$ is the cosine similarity between a fingerprint representation and the corresponding group centroid.
Minimizing $\mathcal{L}^{(i)}_{\mathrm{pr}}$ encourages all fingerprints within each group to be close in the representation space, thereby promoting stable group-level behavior.

\textbf{Group dispersion $\mathcal{L}^{(i)}_{\mathrm{ds}}$.}
We exploit reference classifiers only during fingerprint generation to prevent a fingerprint group from inducing consistent responses on independently trained models. Let $\mathcal{R}$ denote a reference classifier set consisting of independently trained classifiers. At each iteration $t$, we randomly sample a subset $\mathcal{R}_t \subseteq \mathcal{R}$ to compute group dispersion, which mitigates overfitting to a fixed set of reference models.

For a sampled reference classifier $g \in \mathcal{R}_t$ and a fingerprint $\hat{x}_{ij} \in \mathcal{T}_i$, let $q_{ij}^{(g)} \in (0,1)^C$ denote the output probability vector, where $[q_{ij}^{(g)}]_c$ is the predicted probability that an object belonging to the $c$-th semantic class is present in $\hat{x}_{ij}$. To quantify how differently the fingerprints in group $\mathcal{T}_i$ are perceived by the reference classifier $g$, we compute the dispersion score---the average variance of its output probabilities across the $M$ group members---by $\mathrm{Disp}_{i}^{(g)}=\frac{1}{C}\sum_{c=1}^{C}
\biggl(\frac{1}{M}\sum_{j=1}^{M}
\left([q_{ij}^{(g)}]_c-\bar{q}_{i,c}^{(g)}\right)^2
\biggr)$,
where $\bar{q}_{i,c}^{(g)} = \frac{1}{M} \sum_{j=1}^{M} [q_{ij}^{(g)}]_c$ is the average response of classifier $g$ on the $c$-th class within group $\mathcal{T}_i$.
We then define group-dispersion loss at iteration $t$ by
$\mathcal{L}^{(i)}_{\mathrm{ds}}(t)=
\frac{1}{|\mathcal{R}_t|}
\sum_{g \in \mathcal{R}_t}
\mathrm{Disp}_{i}^{(g)}$.
Since $\mathcal{L}^{(i)}_{\mathrm{ds}}(t)$ appears with a negative sign in the overall objective in Eq.~\eqref{eq:overall_objective}, minimizing the total loss encourages larger inner-group dispersion on sampled reference classifiers. Consequently, fingerprints within the same group become less likely to induce aligned responses on independently trained models, thereby improving separability from unrelated model lineages.

\textbf{Perturbation regularization $\mathcal{L}^{(i)}_{\mathrm{rg}}$.}
To keep each fingerprint visually plausible and close to its corresponding sampled input, we introduce a perturbation regularization term $\mathcal{L}^{(i)}_{\mathrm{rg}}$. Specifically, for the $i$-th fingerprint group $\mathcal{T}_i$ with perturbations $\Delta_i=\{\delta_{ij}\}_{j=1}^{M}$, we define 
$\mathcal{L}^{(i)}_{\mathrm{rg}} = \frac{1}{M} \sum_{j=1}^{M} \frac{1}{d} \left\|\delta_{ij}\right\|_1$,
where $d$ is the input dimensionality. Minimizing $\mathcal{L}^{(i)}_{\mathrm{rg}}$ constrains the average magnitude within each group, preventing fingerprints from deviating excessively from the sampled inputs. This regularization preserves the visual plausibility of the generated fingerprints.

\subsection{Ownership Verification}
\label{sec:verify_method}
During ownership verification, the verifier evaluates a suspect model $f_v$, accessible only through black-box queries, using the grouped fingerprint set $\mathcal{T}=\{\mathcal{T}_i\}_{i=1}^{N}$. The verification procedure consists of three steps: task-aware semantic projection, inner-group consistency computation, and decision aggregation. We describe each step in detail below.

\textbf{Task-aware semantic projection.}
Models designed for different tasks produce outputs in heterogeneous formats. When a fingerprint is used to query a suspect model, the returned output may differ in format from that of the protected model due to cross-task model repurposing, making direct comparison infeasible. To address this, we propose a task-aware semantic projection mechanism that maps heterogeneous outputs into a unified representation. The key idea is to preserve information about the presence of semantic classes in each fingerprint response while discarding task-specific structural details that are not comparable across tasks. Specifically, we first define a projection operator $\Pi_{\mathrm{task}}(\cdot)$ that transforms a task-specific output into a binary class-presence vector. For the output $y_{ij}$ obtained by querying the suspect model $f_v$ with fingerprint $\hat{x}_{ij}\in\mathcal{T}_i$, we denote this binary vector by $r_{ij}=\Pi_{\mathrm{task}}(y_{ij})\in\{0,1\}^C$, where $[r_{ij}]_c=1$ if the $c$-th semantic class is present in $y_{ij}$, and $[r_{ij}]_c=0$ otherwise. Then, we normalize $r_{ij}$ to obtain the semantic-presence vector $s_{ij}=\frac{r_{ij}}{\|r_{ij}\|_1}$. This transformation enables outputs from diverse models and tasks to be represented in a unified semantic space, facilitating subsequent consistency analysis.

\textbf{Inner-group consistency computation.}
After obtaining the projected semantic-presence vectors $\{s_{ij}\}_{j=1}^{M}$ for a fingerprint group $\mathcal{T}_i$, we quantify inner-group consistency among the fingerprints in $\mathcal{T}_i$ by measuring pairwise similarity among these responses in the unified semantic space. Specifically, for any two fingerprints indexed by $u,v \in \{1,\ldots,M\}$ with $u<v$, we compute the Jensen-Shannon divergence (JSD)~\citep{lin2002divergence} between their projected semantic-presence vectors and normalize it to obtain a similarity score in $[0,1]$, i.e., $\mathrm{Sim}(s_{iu},s_{iv})
=
1-\min\!\left(\frac{\mathrm{JSD}(s_{iu},s_{iv})}{\ln 2},\,1\right)$.
Then, we define the inner-group consistency score of $\mathcal{T}_i$ as the average pairwise similarity over all unordered fingerprint pairs, i.e., $s_i(f_v)
=
\frac{2}{M(M-1)}
\sum_{1\le u<v\le M}
\mathrm{Sim}(s_{iu},s_{iv})$.
A larger value of $s_i(f_v)$ indicates that the fingerprints in $\mathcal{T}_i$ elicit more consistent semantic responses from the suspect model.

\textbf{Decision aggregation.}
After computing the inner-group consistency score for each group, we utilize these scores to determine the ownership of the suspect model $f_v$. Intuitively, a pirated model derived from the protected model is expected to preserve the group-consistency behavior induced by the fingerprint set, whereas an independently trained model is likely to exhibit inconsistent patterns.

To this end, we aggregate the inner-group consistency scores across all fingerprint groups to obtain a final ownership score, i.e., $S(f_v) =\frac{1}{N} \sum_{i=1}^{N} s_i(f_v)$. This score measures the overall behavioral alignment of the suspect model $f_v$ with the protected model $f^{t_s}$. If $S(f_v)\ge \theta$, where $\theta$ is a predefined decision threshold, the model is classified as pirated; otherwise, it is considered independent.

\section{Experiments}
\label{sec:exp}
\subsection{Experimental Setup}
\label{sec:exp:setup}

\textbf{Datasets and tasks.}
We evaluate UniReFP on three image datasets: Oxford-IIIT Pet (PET) \citep{OxfordPET2012}, PASCAL VOC 2012 (VOC) \citep{VOC2010}, and COCO-Stuff 10K (COCO) \citep{COCO2018}. PET mainly contains single-object scenes, whereas VOC and COCO include multi-object scenes with multiple co-occurring classes. These datasets provide annotations for three representative vision tasks---image classification (CLS), object detection (DET), and semantic segmentation (SEG)---over a shared class-level semantic space. This setting allows us to train task-specific models with heterogeneous output formats while evaluating whether ownership evidence can be preserved under cross-task model repurposing attacks.

\textbf{Model construction for fingerprint generation.} To generate fingerprints for a protected model, the model owner constructs a surrogate classifier and a set of independently trained reference classifiers. We describe detailed experimental settings for constructing these models below.
\begin{itemize}[left=0pt]
     \item \textit{Protected model construction.} For each dataset, we instantiate one protected model for each of the three tasks. More concretely, we use a ResNet-18 \citep{he2016deep} for image classification, a Faster R-CNN \citep{ ren2015faster} for object detection, and a U-Net-based model \citep{ronneberger2015u} for semantic segmentation. 
    
     \item \textit{Surrogate classifier construction.} We use the protected model directly as the surrogate classifier when it is already a classifier, and otherwise construct a surrogate classifier by replacing the original task-specific head with a classifier head while retaining the backbone of the protected model.

     \item \textit{Reference classifier set construction.} For each dataset, we construct a reference classifier set $\mathcal{R}$ for computing the group-dispersion loss. Each set consists of three independently trained classifiers with different architectures---ResNet-18, ResNet-50, and MobileNet---which share the same class-level semantic space as the protected model. To reduce overfitting, at each optimization epoch, we randomly sample a subset $\mathcal{R}_t \subseteq \mathcal{R}$ with $|\mathcal{R}_t|=1$ for loss computation.
     
\end{itemize}

\textbf{Model construction for performance evaluation.}
For each dataset, we construct two disjoint model sets: a pirated set containing models derived from the protected model and an independent set consisting of independently trained models. All these models are unseen during fingerprint generation. Both sets are constructed under both cross-task model repurposing and commonly-studied model modification attack settings, as described below. 
\begin{itemize}[left=0pt]
    \item \textit{Cross-task model repurposing attacks.} For each task pair $(t_s,t_a)$ with $t_a \neq t_s$, we evaluate UniReFP in a cross-task model repurposing setting, where a model trained for task $t_s$ is adapted to the target task $t_a$. 
    The pirated set contains 50 repurposed models for each source-target task pair, while the independent set consists of 50 independently trained target-task models disjoint from the pirated set, constructed with different architectures and training configurations.

    \item \textit{Commonly-studied model modification attacks.} For each task, we evaluate UniReFP in a commonly-studied model modification attack setting, where a protected model is altered to produce pirated variants. The pirated set is constructed by applying five types of attacks to the protected model—fine-tuning (FT), pruning (PR), knowledge distillation (KD), prune–fine-tuning (PFT), and noise–fine-tuning (NFT). Each attack generates 10 pirated models with different random seeds, resulting in a total of 50 pirated models. The independent set consists of 50 independently trained models for the same task, built with different architectures and training configurations.

\end{itemize}

\textbf{The evaluation metric.}
We use the Area Under the Receiver Operating Characteristic Curve (AUC) as the primary evaluation metric. AUC measures the probability that a randomly selected pirated model receives a higher verification score than a randomly selected independently trained model, thus quantifying the discriminative power of the method across all decision thresholds. We report the mean and standard deviation over five runs to account for variability across fingerprint generation.

\textbf{Baselines.}
We compare UniReFP with five representative fingerprinting approaches, including three classification-oriented approaches (AnaFP~\cite{yang26anafp}, AKH~\cite{QURD25}, and TFP~\cite{TMFA22}) and two task-agnostic approaches (MetaV~\cite{MVAM22} and LiteGuard~\cite{yang26liteguard}). AnaFP, AKH, and TFP are designed for classification models and rely on classification outputs for both fingerprint generation and verification, while MetaV and LiteGuard can be applied to more tasks, such as classification and segmentation, but require fingerprint generation and verification to operate under the same output format. Hence, no existing fingerprinting method directly supports the assessment of UniReFP under cross-task model repurposing attacks. To enable comparison in this setting, we adapt from TFP to construct a baseline called TFP-M by retaining TFP’s original fingerprint generation procedure while replacing its verification mechanism with UniReFP’s task-unified verification protocol. As TFP is designed for classification tasks, TFP-M is only applicable to scenarios where a classifier is repurposed for object detection or semantic segmentation, instead of the case that the protected model is originally a detection or segmentation model.

\textbf{Other implementation details.} To evaluate UniReFP, we set $M$=10 and $N$=10, yielding 100 fingerprints in total. For fair comparison, the same number of fingerprints is used for all baseline methods. Besides, we set $\lambda$=0.001, $p_{\mathrm{lo}}=0.1$, $p_{\mathrm{hi}}=0.9$, $\theta = 0.5$ and $|\mathcal{C}^{\mathrm{sel}}|$=5.

\subsection{The Effectiveness of UniReFP}
\label{sec:exp:taskconv}

We assess the effectiveness of UniReFP by considering three attack settings: 1) only cross-task model repurposing attacks, 2) only commonly-studied model modification attacks, and 3) combined cross-task repurposing and commonly-studied modification attacks.

\textbf{Cross-task Model Repurposing Attacks.}
Table~\ref{tab:cross_task_repurpose} illustrates the superior effectiveness of UniReFP by reporting the AUCs by UniReFP and TFP-M across all repurposing cases among three tasks (i.e., CLS, DET, and SEG) on different datasets. We can observe that UniReFP achieves strong performance across datasets and repurposing cases, validating UniReFP's effectiveness. Specifically, among the 18 repurposing cases, all AUCs are above 0.8, and 13 settings reach at least 0.97. For instance, UniReFP achieves near-perfect performance in CLS$\rightarrow$SEG and CLS$\rightarrow$DET on all datasets, DET$\rightarrow$SEG on PET/COCO, and SEG$\rightarrow$DET on VOC/COCO. Spanning different protected and repurposed models, this shows that UniReFP can reliably distinguish repurposed models from independently trained models even when the suspect model returns an output with a different format from the protected model. TFP-M is only applicable where $t_s$ is the classification task. As shown in Table~\ref{tab:cross_task_repurpose}, TFP-M consistently achieves lower AUCs than UniReFP among different repurposed tasks across different datasets.

\begin{table*}[t]
\setlength{\tabcolsep}{1.8pt}
\centering
\caption{AUCs under cross-task model repurposing attacks across datasets. $t_s$ and $t_a$ represent the tasks for protected and repurposed models, respectively.
}
\resizebox{\textwidth}{!}{%
\begin{tabular}{ll|ccc|ccc|ccc}
\toprule

& & \multicolumn{3}{c|}{\makecell{\textbf{PET}}}
 & \multicolumn{3}{c|}{\makecell{\textbf{VOC}}}
 & \multicolumn{3}{c}{\makecell{\textbf{COCO}}} \\

\midrule
\multicolumn{2}{c|}{\diagbox{\makecell{{\large $t_s$}}}
{\makecell{{\large $t_a$}}}}
& CLS & DET & SEG
& CLS & DET & SEG
& CLS & DET & SEG\\

\midrule
\multirow{3}{*}{\textbf{UniReFP}} &\makecell{CLS}
& -- & 1.000$\pm$0.000 & 1.000$\pm$0.000
& -- & 0.993$\pm$0.012 & 1.000$\pm$0.000
& -- & 0.988$\pm$0.012 & 1.000$\pm$0.000 \\

&\makecell{DET}
& 0.895$\pm$0.066 & -- & 0.974$\pm$0.037
& 0.973$\pm$0.038 & -- & 0.841$\pm$0.017
& 0.868$\pm$0.172 & -- & 1.000$\pm$0.000 \\

&\makecell{SEG}
& 0.978$\pm$0.016 & 0.804$\pm$0.013 & --
& 1.000$\pm$0.000 & 1.000$\pm$0.000 & --
& 0.912$\pm$0.040 & 0.993$\pm$0.010 & -- \\

\midrule

\multirow{3}{*}{TFP-M} 
&\makecell{CLS}
& -- & 0.770$\pm$0.095 & 0.901$\pm$0.019
& -- & 0.916$\pm$0.040 & 0.928$\pm$0.021
& -- & 0.912$\pm$0.034 & 0.945$\pm$0.013 \\

&\makecell{DET}
& -- & -- & --
& -- & -- & --
& -- & -- & -- \\

&\makecell{SEG}
& -- & -- & --
& -- & -- & --
& -- & -- & -- \\

\bottomrule
\end{tabular}%
}
\label{tab:cross_task_repurpose}
\end{table*}

\textbf{Commonly-studied model modification attacks.}
Table~\ref{tab:intask_performance} shows the results achieved by UniReFP across three tasks on three datasets under commonly-studied model modification attacks. From the table, we can see that UniReFP maintains strong and reliable discriminative capability under commonly-studied model modification attacks, as AUCs range from 0.817 to 0.957 across all tasks among different datasets. For classification, UniReFP obtains AUCs of 0.889, 0.912, and 0.896 on PET, VOC, and COCO, respectively. For semantic segmentation, UniReFP achieves AUCs of 0.817, 0.957, and 0.893 on the same three datasets. These results show that UniReFP achieves competitive performance compared with representative classification-oriented baselines and task-agnostic baselines. For object detection, where existing fingerprinting methods are not directly applicable, we report UniReFP as the primary method. UniReFP remains effective for object detection, achieving AUCs of 0.875, 0.842, and 0.942 on PET, VOC, and COCO, respectively. These results indicate that UniReFP is not only robust against cross-task repurposing attacks, but also commonly-studied model modification attacks.

\begin{table*}[t]
\setlength{\tabcolsep}{1.8pt}
\centering
\caption{AUCs under the commonly-studied model modification attacks on three datasets.}
\resizebox{\textwidth}{!}{%
\begin{tabular}{l|ccc|ccc|ccc}
\toprule
& \multicolumn{3}{c|}{\makecell{\textbf{CLS}}}
& \multicolumn{3}{c|}{\makecell{\textbf{DET}}}
& \multicolumn{3}{c}{\makecell{\textbf{SEG}}} \\
\cmidrule(lr){2-4}\cmidrule(lr){5-7}\cmidrule(lr){8-10}
& \textbf{PET} & \textbf{VOC} & \textbf{COCO}
& \textbf{PET} & \textbf{VOC} & \textbf{COCO}
& \textbf{PET} & \textbf{VOC} & \textbf{COCO} \\
\midrule
\textbf{UniReFP}
& 0.889$\pm$0.008 & 0.912$\pm$0.009 & 0.896$\pm$0.018
& \textbf{0.875$\pm$0.023} & \textbf{0.842$\pm$0.073} & \textbf{0.942$\pm$0.026}
& {0.817$\pm$0.010} & {0.957$\pm$0.009} & \textbf{0.893$\pm$0.006} \\
TFP
& 0.881$\pm$0.004 & 0.886$\pm$0.010 & 0.912$\pm$0.016
& -- & -- & --
& -- & -- & -- \\
MetaV
& 0.754$\pm$0.058 & 0.871$\pm$0.011 & 0.834$\pm$0.006
& -- & -- & --
& 0.809$\pm$0.031 & 0.899$\pm$0.016 & 0.646$\pm$0.043 \\
{LiteGuard}
& {0.920$\pm$0.018} & \textbf{0.957$\pm$0.013} & \textbf{0.941$\pm$0.006}
& -- & -- & --
& \textbf{0.884$\pm$0.047} & \textbf{0.986$\pm$0.005} & 0.793$\pm$0.013 \\
AKH
& 0.861$\pm$0.024 & -- & --
& -- & -- & --
& -- & -- & -- \\
AnaFP
& \textbf{0.945$\pm$0.016} & -- & --
& -- & -- & --
& -- & -- & -- \\
\bottomrule
\end{tabular}%
}
\label{tab:intask_performance}
\vspace{-5pt}
\end{table*}

\textbf{Combined cross-task repurposing and commonly-studied modification attacks.}
This setting allows us to assess whether UniReFP remains effective in distinguishing the pirated models from independently trained ones under more challenging and realistic attack scenarios. Specifically, we conduct the assessment on the COCO dataset across six ordered source-target task pairs among CLS, DET, and SEG. For each task pair, we evaluate five attack combinations: BP$\rightarrow$RP, RP$\rightarrow$NFT, RP$\rightarrow$PFT, BP$\rightarrow$RP$\rightarrow$NFT, and BP$\rightarrow$RP$\rightarrow$PFT, where BP, RP, NFT, and PFT denote backbone pruning, cross-task model repurposing, noise-fine-tuning, and prune-fine-tuning, respectively. These combinations cover three representative cases: modifying the backbone before repurposing, modifying the repurposed model after repurposing, and modifying the model at both stages.

\begin{wraptable}{r}{0.72\textwidth}
\centering
\caption{AUCs under combined repurposing and modification attacks.}
\label{tab:combined_attack}
\small
\setlength{\tabcolsep}{3pt}
\renewcommand{\arraystretch}{1.15}
\resizebox{0.72\textwidth}{!}{%
\begin{tabular}{c|ccccc}
\toprule
\textbf{{\large $t_s$} $\rightarrow$ {\large $t_a$}}
& \makecell{\textbf{BP}$\rightarrow$\textbf{RP}}
& \makecell{\textbf{RP}$\rightarrow$\textbf{NFT}}
& \makecell{\textbf{RP}$\rightarrow$\textbf{PFT}}
& \makecell{\textbf{BP}$\rightarrow$\textbf{RP}\\$\rightarrow$\textbf{NFT}}
& \makecell{\textbf{BP}$\rightarrow$\textbf{RP}\\$\rightarrow$\textbf{PFT}} \\
\midrule
CLS$\rightarrow$DET & 0.872$\pm$0.024 & 0.932$\pm$0.011 & 0.956$\pm$0.015 & 0.859$\pm$0.028 & 0.866$\pm$0.026 \\
CLS$\rightarrow$SEG & 0.965$\pm$0.018 & 0.997$\pm$0.005 & 0.998$\pm$0.004 & 0.952$\pm$0.023 & 0.961$\pm$0.020 \\
DET$\rightarrow$CLS & 0.788$\pm$0.126 & 0.904$\pm$0.134 & 0.813$\pm$0.120 & 0.758$\pm$0.099 & 0.803$\pm$0.065 \\
DET$\rightarrow$SEG & 0.939$\pm$0.071 & 0.998$\pm$0.002 & 1.000$\pm$0.000 & 0.978$\pm$0.032 & 0.975$\pm$0.005 \\
SEG$\rightarrow$CLS & 0.753$\pm$0.053 & 0.833$\pm$0.077 & 0.852$\pm$0.032 & 0.738$\pm$0.033 & 0.735$\pm$0.075 \\
SEG$\rightarrow$DET & 0.789$\pm$0.070 & 0.935$\pm$0.103 & 0.967$\pm$0.053 & 0.742$\pm$0.016 & 0.976$\pm$0.025 \\
\bottomrule
\end{tabular}%
}
\end{wraptable}

Experimental results are summarized in Table~\ref{tab:combined_attack}. As demonstrated, UniReFP remains strong discriminative capability under these combined attacks, as most of AUCs are large, and particularly AUCs in some cases approach 1.0 (i.e., near-perfect discriminative capability). Compared with the repurposing-only setting, the AUCs decrease in several cases, especially when backbone pruning is applied before cross-task repurposing. This finding indicates that the combined attack makes ownership verification more challenging.

\subsection{Robustness to Commonly-studied Model Modification Attacks}
We evaluate UniReFP's robustness to commonly-studied model modification attacks using five representative attacks: fine-tuning (FT), pruning (PR), knowledge distillation (KD), pruning-fine-tuning (PFT), and noise-fine-tuning (NFT). Table~\ref{tab:model_modification_attacks_all_source_tasks} reports the AUCs of UniReFP and five baselines across three datasets and three tasks under these attacks. As shown, UniReFP consistently achieves strong robustness across different tasks and datasets. In particular, its AUC remains stable under FT, PR, PFT, and NFT for classification, object detection, and semantic segmentation. This indicates that the fingerprint-induced inner-group behavior captured by UniReFP is largely preserved after these modifications. It is noteworthy that KD is the most challenging attack, as teacher-student transfer can significantly alter fingerprint-specific response behavior, consequently resulting in severe degradation in robustness. Nevertheless, UniReFP remains competitive with the baselines under KD. These results highlight that defending against KD successfully is still an open research challenge.

\begin{table*}[t]
\centering
\caption{AUCs under five commonly-studied model modifications across different datasets and tasks.}
\resizebox{\textwidth}{!}{%
\begin{tabular}{lll|ccccc}
\toprule
\textbf{Task} & \textbf{Dataset} & \textbf{Method} & \textbf{FT} & \textbf{PR} & \textbf{KD} & \textbf{NFT} & \textbf{PFT} \\
\midrule

\multirow{14}{*}{\textbf{CLS}}
& \multirow{6}{*}{\textbf{PET}}
& \textbf{UniReFP}    & 1.000$\pm$0.000 & 1.000$\pm$0.000 & 0.445$\pm$0.041 & 1.000$\pm$0.000 & 1.000$\pm$0.000 \\
&& TFP               & 1.000$\pm$0.000 & 1.000$\pm$0.000 & 0.409$\pm$0.015 & 1.000$\pm$0.000 & 0.996$\pm$0.004 \\
&& MetaV             & 0.819$\pm$0.097 & 0.921$\pm$0.032 & 0.364$\pm$0.115 & 0.787$\pm$0.143 & 0.879$\pm$0.090 \\
&& LiteGuard         & 1.000$\pm$0.000 & 1.000$\pm$0.000 & 0.598$\pm$0.094 & 1.000$\pm$0.000 & 1.000$\pm$0.000 \\
&& AKH               & 1.000$\pm$0.000 & 1.000$\pm$0.000 & 0.454$\pm$0.047 & 0.983$\pm$0.029 & 0.867$\pm$0.057 \\
&& AnaFP             & 1.000$\pm$0.000 & 1.000$\pm$0.000 & 0.727$\pm$0.081 & 1.000$\pm$0.000 & 1.000$\pm$0.000 \\
\cmidrule(lr){2-8}

& \multirow{4}{*}{\textbf{VOC}}
& \textbf{UniReFP}    & 1.000$\pm$0.000 & 1.000$\pm$0.000 & 0.559$\pm$0.042 & 1.000$\pm$0.000 & 0.997$\pm$0.001 \\
&& TFP               & 1.000$\pm$0.000 & 1.000$\pm$0.000 & 0.460$\pm$0.049 & 1.000$\pm$0.000 & 0.968$\pm$0.012 \\
&& MetaV             & 0.991$\pm$0.003 & 0.968$\pm$0.041 & 0.519$\pm$0.032 & 0.976$\pm$0.010 & 0.901$\pm$0.021 \\
&& LiteGuard         & 1.000$\pm$0.000 & 1.000$\pm$0.000 & 0.785$\pm$0.064 & 1.000$\pm$0.000 & 1.000$\pm$0.000 \\
\cmidrule(lr){2-8}

& \multirow{4}{*}{\textbf{COCO}}
& \textbf{UniReFP}    & 1.000$\pm$0.000 & 1.000$\pm$0.000 & 0.478$\pm$0.089 & 1.000$\pm$0.000 & 1.000$\pm$0.000 \\
&& TFP               & 1.000$\pm$0.000 & 1.000$\pm$0.000 & 0.630$\pm$0.077 & 1.000$\pm$0.000 & 0.928$\pm$0.014 \\
&& MetaV             & 0.931$\pm$0.022 & 1.000$\pm$0.000 & 0.385$\pm$0.000 & 0.987$\pm$0.009 & 0.867$\pm$0.018 \\
&& LiteGuard         & 1.000$\pm$0.000 & 1.000$\pm$0.000 & 0.705$\pm$0.029 & 1.000$\pm$0.000 & 1.000$\pm$0.000 \\
\midrule

\multirow{3}{*}{\textbf{DET}}
& \textbf{PET}
& \textbf{UniReFP}    & 1.000$\pm$0.000 & 0.889$\pm$0.000 & 0.518$\pm$0.092 & 1.000$\pm$0.000 & 0.990$\pm$0.014 \\
\cmidrule(lr){2-8}
& \textbf{VOC}
& \textbf{UniReFP}    & 0.821$\pm$0.119 & 1.000$\pm$0.000 & 0.792$\pm$0.034 & 0.804$\pm$0.119 & 0.774$\pm$0.119 \\
\cmidrule(lr){2-8}
& \textbf{COCO}
& \textbf{UniReFP}    & 1.000$\pm$0.000 & 0.889$\pm$0.000 & 0.642$\pm$0.216 & 1.000$\pm$0.000 & 1.000$\pm$0.000 \\
\midrule

\multirow{9}{*}{\textbf{SEG}}
& \multirow{3}{*}{\textbf{PET}}
& \textbf{UniReFP}    & 1.000$\pm$0.000 & 0.985$\pm$0.006 & 0.260$\pm$0.034 & 0.926$\pm$0.017 & 0.915$\pm$0.027 \\
&& MetaV              & 0.896$\pm$0.091 & 0.837$\pm$0.110 & 0.476$\pm$0.115 & 0.863$\pm$0.050 & 0.874$\pm$0.051 \\
&& LiteGuard          & 1.000$\pm$0.000 & 0.951$\pm$0.042 & 0.636$\pm$0.048 & 0.844$\pm$0.137 & 0.993$\pm$0.009 \\
\cmidrule(lr){2-8}

& \multirow{3}{*}{\textbf{VOC}}
& \textbf{UniReFP}    & 1.000$\pm$0.000 & 1.000$\pm$0.000 & 0.662$\pm$0.069 & 1.000$\pm$0.000 & 1.000$\pm$0.000 \\
&& MetaV             & 0.951$\pm$0.030 & 0.840$\pm$0.049 & 0.720$\pm$0.093 & 0.929$\pm$0.011 & 0.933$\pm$0.014 \\
&& LiteGuard         & 1.000$\pm$0.000 & 0.978$\pm$0.006 & 0.911$\pm$0.037 & 1.000$\pm$0.000 & 1.000$\pm$0.000 \\
\cmidrule(lr){2-8}

& \multirow{3}{*}{\textbf{COCO}}
& \textbf{UniReFP}    & 0.833$\pm$0.003 & 0.953$\pm$0.021 & 0.998$\pm$0.003 & 0.838$\pm$0.006 & 0.836$\pm$0.005 \\
&& MetaV             & 0.779$\pm$0.083 & 0.516$\pm$0.097 & 0.439$\pm$0.243 & 0.712$\pm$0.062 & 0.800$\pm$0.034 \\
&& LiteGuard         & 1.000$\pm$0.000 & 0.690$\pm$0.017 & 0.285$\pm$0.049 & 1.000$\pm$0.000 & 1.000$\pm$0.000 \\
\bottomrule
\end{tabular}%
}

\label{tab:model_modification_attacks_all_source_tasks}
\end{table*}

\section{Conclusions}

In this paper, we study the problem of realizing robust model fingerprinting for vision models against cross-task repurposing attacks that can render fingerprints generated by existing approaches inapplicable, as the repurposed models have heterogeneous output formats. To address this issue, we develop UniReFP, a unified model fingerprinting framework applicable to diverse vision models across different tasks. The effectiveness of UniReFP is enabled by three innovative design components: (i) constructing a surrogate classifier from a protected model to enable a unified fingerprinting paradigm, (ii) optimizing grouped fingerprints by jointly considering semantic- and representation-level group consistency, group dispersion on independently trained classifiers, and perturbation regularization, and (iii) designing a task-aware semantic projection mechanism to map heterogeneous outputs into a unified representation, enabling cross-task black-box verification. Extensive experiments demonstrate the effectiveness of UniReFP, consistently achieving strong ownership verification performance under both cross-task repurposing attacks and commonly studied model modification attacks.

\section*{Acknowledgments and Disclosures}

The work of X. Fu was supported in part by the U.S. National Science Foundation under Grants CNS-2107057 and SSF-2504152. The work of L. Yao and M. Pan was supported in part by the U.S. National Science Foundation under Grants CNS-2318664, CSR-2403249, and CNS-2431596.
The authors declare no competing interests.

\newpage

\bibliographystyle{plainnat}
\bibliography{ref}

\clearpage
\appendix

\section{Additional Experimental Results}
\label{sec:appendix}

\subsection{Impact of Hyperparameters}
\label{sec:appendix_impactparameters}

In this section, we analyze the impact of hyperparameter choices on UniReFP's performance. We focus on the setting where the protected model is trained for image classification, and evaluate UniReFP under two attack scenarios. First, under cross-task model repurposing attacks, the adversary repurposes the protected model to two target tasks: object detection and semantic segmentation. Second, under commonly-studied model modification attacks, the protected model remains a classifier with the original output format, but is modified using commonly-studied attack operations. Accordingly, in the following analysis, classification results correspond to commonly-studied modifications of the protected classifier without model repurposing, whereas detection and segmentation results correspond to models repurposed from the protected classifier to the respective target tasks.

\subsubsection{The impact of the reference classifier set}

UniReFP leverages a set of independently trained reference classifiers during fingerprint generation to improve uniqueness (Sec.~\ref{sec:method}).
Intuitively, a larger and more diverse reference set provides stronger negative evidence, discouraging fingerprints from inducing high inner-group consistency on independently trained models. To study the impact of the reference classifier set, we vary the set size from $1$ to $4$ while keeping all other hyperparameters fixed. For each setting, we generate fingerprints using the corresponding reference set and evaluate AUC under the same protocol as in Sec.~\ref{sec:exp:setup}. 

\begin{figure*}[htbp]
    \centering
    \includegraphics[width=0.9\linewidth]{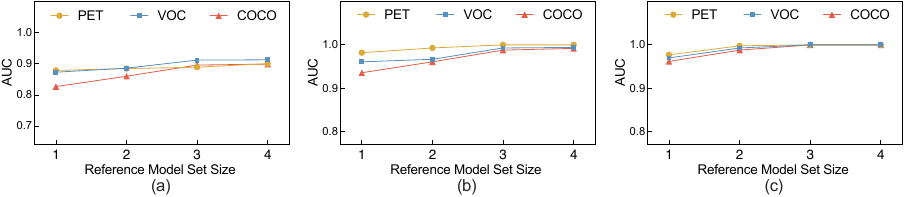}
    \caption{AUCs for different sizes of reference classifier set under different tasks: (a) image classification, (b) object detection, and (c) semantic segmentation.}
    \label{fig:abl_modelsize}
\end{figure*}

Figure~\ref{fig:abl_modelsize} reports the experimental results on PET, VOC, and COCO across three tasks. As illustrated, increasing the size of the reference classifier set consistently improves AUC across datasets and tasks, with diminishing returns beyond three classifiers. For image classification (Fig.~\ref{fig:abl_modelsize}(a)), AUC increases monotonically with set size. The gain is most pronounced on the dataset COCO, increasing from $0.827$ (size $1$) to $0.896$ (size $3$), suggesting that multi-object scenes increase the chance of inner-group consistency when uniqueness constraints are weak.
For object detection (Fig.~\ref{fig:abl_modelsize}(b)) and semantic segmentation (Fig.~\ref{fig:abl_modelsize}(c)), AUC is already high with a small reference set, and quickly saturates: by size $3$, the detection task reaches the AUC of $0.988$ (COCO) and $0.993$ (VOC), while the segmentation task achieves the AUC of $1.000$ on VOC and COCO and remains near-perfect on PET.
The results show that a relatively small reference set is sufficient for strong performance, incurring modest computational cost of constructing the reference classifier set. In practice, publicly available models can serve as reference classifiers, offering a cost-effective alternative to training from scratch.

\subsubsection{The impact of the total number of fingerprints}
\label{sec:ablation:fpnum}
\begin{figure*}[htbp]
    \centering
    \includegraphics[width=0.9\linewidth]{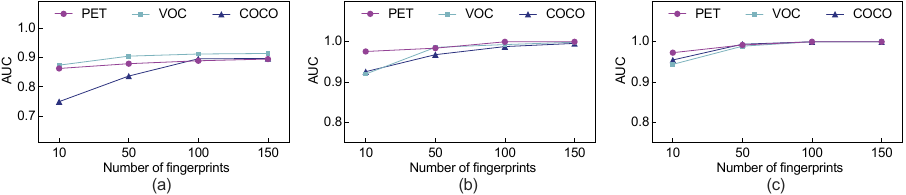}
    \caption{AUCs for different numbers of fingerprints under different tasks: (a) image classification, (b) object detection, and (c) semantic segmentation.}
    \label{fig:abl_query_num}
\end{figure*}

We study the impact of varying the total number of fingerprints. The results are summarized in Fig.~\ref{fig:abl_query_num}. Across all datasets and tasks, increasing the number of fingerprints consistently improves AUC, with clear diminishing returns beyond 100 fingerprints. For instance, for image classification (Fig.~\ref{fig:abl_query_num}(a)), AUC increases steadily as the number of fingerprints grows. The most pronounced gain is observed on the dataset COCO, where AUC rises from 0.750 at 10 fingerprints to 0.896 at 100 fingerprints. For semantic segmentation (Fig.~\ref{fig:abl_query_num}(c)), AUC is already high with a small number of fingerprints and quickly stabilizes: at 50 fingerprints, AUC reaches 0.994 (COCO), 0.989 (VOC), and 0.992 (PET), and achieves 1.000 at 100 fingerprints.
These results indicate that UniReFP achieves strong discriminability with a moderate number of fingerprints. Accordingly, we use $100$ fingerprints as the default setting in this work.

\begin{figure*}[htbp]
    \centering
    \includegraphics[width=0.9\linewidth]{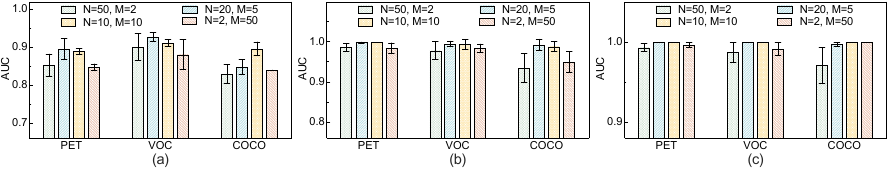}
    \caption{
    AUCs under different fingerprint groups $N$ and group members $M$ with fixed fingerprint number $Q=NM=100$ across datasets and tasks: (a) image classification, (b) object detection, and (c) semantic segmentation.}
    \label{fig:abl_MN_all}
\end{figure*}

\subsubsection{The impact of the combination of M and N}
\label{sec:ablation:MN}
We study how the $(N,M)$ split affects verification with the fixed fingerprint number $Q=100$, as we partition the fingerprints into $N$ groups with each containing $M$ fingerprints. Experimental results are summarized in Fig.~\ref{fig:abl_MN_all}. As shown, the performance is generally stronger when the query budget is distributed into a moderate number of groups with a moderate group size, rather than using an extreme split. In particular, the setting $(2,50)$ often underperforms because too few groups limit verification reliability. For example, compared with the balanced setting $(10,10)$, $(2,50)$ reduces the AUC from $0.896$ to $0.839$ under the classification task, and from $0.988$ to $0.950$ under the classification-to-detection repurposing attacks setting, on the COCO dataset.
Meanwhile, $M$ should not be too small, since estimating inner-group consistency from only a few fingerprints can be unreliable. This can be observed by comparing $(50,2)$ with the more balanced setting $(10,10)$. For example, the AUC increases from $0.829$ to $0.896$ under the classification task, from $0.934$ to $0.988$ under the classification-to-detection repurposing setting, and from $0.971$ to $1.000$ under the classification-to-segmentation repurposing setting, on the COCO dataset. These results indicate that a very small group size may weaken the reliability of group-consistency estimation.
Overall, UniReFP is not overly sensitive to the exact $(N,M)$ split as long as both the number of groups and the group size remain moderate.

\begin{figure*}[htbp]
    \centering
    \includegraphics[width=0.9\linewidth]{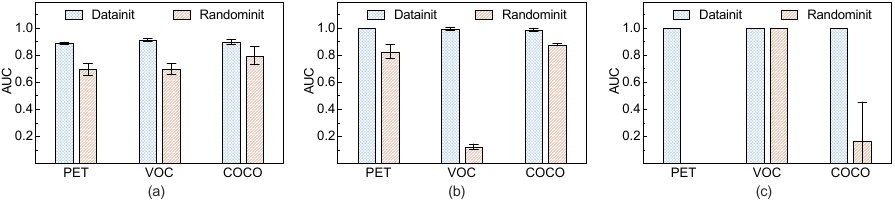}
    \caption{AUCs for different fingerprint initialization methods: natural data samples (Datainit) versus random noise (Randominit), evaluated under different tasks: (a) image classification, (b) object detection, and (c) semantic segmentation.}
    \label{fig:abl_initialization}
\end{figure*}
\subsubsection{The impact of fingerprint initialization method}
\label{sec:ablation:init}

We study the impact of fingerprint initialization on verification performance using two strategies: DataInit, which initializes each fingerprint from a natural sample with a small optimized perturbation, and RandomInit, which starts from random noise. As shown in Figure~\ref{fig:abl_initialization}, DataInit consistently achieves stronger and more stable AUCs across datasets and tasks, whereas RandomInit is highly unstable. For example, when the protected model is repurposed to object detection task, DataInit achieves near-perfect AUCs of $1.000$ (PET), $0.993$ (VOC), and $0.988$ (COCO), while RandomInit drops to $0.827$ (PET), $0.126$ (VOC) and $0.877$ (COCO), respectively. These results suggest that initializing fingerprints from natural samples anchors them to the data manifold, leading to more stable group-wise behavior under cross-task repurposing and improved separability across models.

\subsection{Time Cost of  Constructing Fingerprints}
\label{sec:ablation:runtime}
We evaluate the runtime of generating fingerprints for protected models trained on the image classification task across three datasets. The results are summarized in Table~\ref{tab:runtime_comparison}. As shown, UniReFP incurs relatively low time overhead compared to five representative baselines (TFP, AKH, AnaFP, MetaV, and LiteGuard), and is notably more efficient than optimization-heavy methods such as TFP, AKH, and AnaFP. On the multi-object datasets, UniReFP requires 116s on COCO, outperforming TFP (252s) and LiteGuard (450s), and approaching the efficiency of MetaV (53s). On VOC, UniReFP takes 117s, again faster than TFP (275s) and LiteGuard (452s), and close to MetaV (56s). On PET, UniReFP further reduces runtime to 44s, significantly outperforming TFP (213s), LiteGuard (216s), AKH (122s), and AnaFP (11249s). These results indicate that UniReFP achieves a favorable balance between efficiency and effectiveness. 

\begin{table*}[t]
\centering
\caption{Time overhead of constructing fingerprints for protected models trained on image classification across three datasets.}
\begin{tabular}{l|cccccc}
\toprule
\textbf{Dataset} & \textbf{UniReFP} & \textbf{TFP} & \textbf{MetaV} & \textbf{LiteGuard} & \textbf{AKH} & \textbf{AnaFP} \\
\midrule
COCO & 116s & 275s & 56s & 452s & --   & -- \\
VOC  & 117s & 252s & 53s & 450s & --   & -- \\
PET  & 44s  & 213s & 10s & 216s & 122s & 11249s \\
\bottomrule
\end{tabular}

\label{tab:runtime_comparison}
\end{table*}


\section{Additional Implementation Details}
\label{sec:appendix_implementation_details}

\subsection{Additional Details for Fingerprint Generation}
For surrogate classifier construction, we freeze the backbone and train the classifier head for 10 epochs using AdamW optimizer with a learning rate of $10^{-3}$, weight decay $10^{-4}$, batch size of 16, and a cosine learning-rate scheduler with minimum learning rate $10^{-5}$. For fingerprint optimization, we optimize the trainable perturbations for 100 epochs using AdamW optimizer with a learning rate of $10^{-2}$.

\subsection{Commonly-studied Model Modification Attacks}
We evaluate five commonly-studied model modification attacks: fine-tuning (FT), pruning (PR), knowledge distillation (KD), noise-fine-tuning (NFT), and prune-fine-tuning (PFT).

\begin{itemize}[left=0pt]
    \item Fine-tuning (FT): We fine-tune the protected model on the original task using the corresponding task-specific objective, while keeping the model structure unchanged.
    \item Pruning (PR): We perform global unstructured pruning at sparsity levels from 10\% to 90\%, increasing by 10\% each time.
    \item Knowledge Distillation (KD): Training a student model, which may use either the same or a different architecture, to match the protected model’s output distribution using KL divergence with temperature set to 1.
    \item Noise-fine-tuning (NFT): We first inject Gaussian noise into each trainable parameter tensor, where the noise magnitude is scaled by the tensor's standard deviation, i.e., $param+= \alpha \cdot std(param) \cdot \mathcal{N}(0,1)$, and then fine-tune the perturbed model on the original task.
    \item Prune-fine-tuning (PFT): We apply global unstructured pruning at a sparsity level of 70\%, followed by fine-tuning to recover task performance.
\end{itemize}

\subsection{Combined Repurposing and Modification Attacks}
We further consider a stronger adaptive setting where cross-task model repurposing is combined with additional model-modification operations. Specifically, we consider five combined attack pipelines: backbone pruning followed by repurposing (BP$\rightarrow$RP), repurposing followed by noise-fine-tuning (RP$\rightarrow$NFT), repurposing followed by prune-fine-tuning (RP$\rightarrow$PFT), backbone pruning followed by repurposing and noise-fine-tuning (BP$\rightarrow$RP$\rightarrow$NFT), and backbone pruning followed by repurposing and prune-fine-tuning (BP$\rightarrow$RP$\rightarrow$PFT).

\begin{itemize}[left=0pt]
    \item BP$\rightarrow$RP: We first apply global unstructured pruning to the backbone of the protected model at a sparsity level of 70\%, and then reuse the pruned backbone to construct a target-task model through cross-task repurposing.

    \item RP$\rightarrow$NFT: We first repurpose the protected model to the target task by replacing the original task-specific prediction module with a target-task module, and then inject Gaussian noise into the repurposed model followed by fine-tuning.

    \item RP$\rightarrow$PFT: We first perform cross-task repurposing to obtain a target-task model, and then apply global unstructured pruning at a sparsity level of 50\% to the repurposed model followed by fine-tuning to recover task performance.

    \item BP$\rightarrow$RP$\rightarrow$NFT: We first prune the protected backbone, use the pruned backbone for cross-task repurposing, and then further apply noise-fine-tuning to the resulting repurposed model.

    \item BP$\rightarrow$RP$\rightarrow$PFT: We first prune the protected backbone, repurpose it to the target task, and then apply prune-fine-tuning to the final repurposed model.
\end{itemize}

\subsection{Instantiations of $\Pi_{\mathrm{task}}$ for Different Tasks}

We instantiate $\Pi_{\mathrm{task}}$ for several representative tasks. Across these tasks, the projection follows a unified principle: we first construct a class-presence vector indicating which semantic classes appear in the predicted output, and then apply $\ell_1$-normalization to obtain a semantic-presence vector $\mathbf{s}_{ij}$. Below is the detailed description of the instantiations.

\textbf{\emph{Image classification.}}
If $f_v$ is a classifier, its output on $\hat{x}_{ij}$ is a label set
$\mathcal{A}_{ij}\subseteq\{1,2,\ldots,C\}$.
We first define a class-presence vector $\mathbf{u}_{ij}\in\{0,1\}^C$ by
\begin{equation}
[\mathbf{u}_{ij}]_c
=
\mathbb{I}\!\left[c\in \mathcal{A}_{ij}\right],
\qquad c\in\{1, 2, \ldots,C\},
\label{eq:cls_presence}
\end{equation}
where $[\mathbf{u}_{ij}]_c$ indicates whether class $c$ is predicted by the classifier. Then, we obtain a semantic-presence vector by $\mathbf{s}_{ij}
=
\Pi_{\mathrm{cls}}(y_{ij})
=
\frac{\mathbf{u}_{ij}}{\|\mathbf{u}_{ij}\|_1}$.

\textbf{\emph{Object detection.}}
If $f_v$ is a detector, its output on $\hat{x}_{ij}$ is a set of predicted instances
$y_{ij}=\{(b_k,c_k)\}_{k=1}^{N_{ij}}$,
where $b_k$ is the predicted bounding box, $c_k\in\{1, 2, \ldots,C\}$ is the predicted class, and $N_{ij}$ is the number of detections.
We map this variable-length output to a binary class-presence vector $\mathbf{u}_{ij}\in\{0,1\}^C$ by
\begin{equation}
[\mathbf{u}_{ij}]_c
=
\mathbb{I}\!\left[\exists\, k \in \{1, 2, \ldots,N_{ij}\}\ \text{, s.t.}\ c_k=c\right],
\qquad c\in\{1, 2, \ldots,C\},
\label{eq:det_presence}
\end{equation}
which indicates whether class $c$ appears among the predicted instances.
Then, we find a semantic-presence vector by $\mathbf{s}_{ij}
=
\Pi_{\mathrm{det}}(y_{ij})
=
\frac{\mathbf{u}_{ij}}{\|\mathbf{u}_{ij}\|_1}$.
This projection discards instance counts and spatial information, retaining only class-presence information in the output.

\textbf{\emph{Semantic segmentation.}}
If $f_v$ is a segmenter, its output on $\hat{x}_{ij}$ is a discrete label mask
$y_{ij}\in\{1, 2, \ldots,C\}^{H\times W}$.
We construct a class-level presence vector $\mathbf{u}_{ij}\in\{0,1\}^C$ by checking whether each semantic class appears anywhere in the predicted mask:
\begin{equation}
[\mathbf{u}_{ij}]_c
=
\mathbb{I}\!\left[\exists\,(h,w)\ \text{s.t.}\ y_{ij}[h,w]=c\right],
\qquad c\in\{1, 2,\ldots,C\},
\label{eq:seg_presence}
\end{equation}
where $(h,w)$ is an index of a pixel located in the $H\times W$ mask, and $\mathbb{I}[\cdot]$ is the indicator function.
Then, we get a semantic-presence vector by $\mathbf{s}_{ij}
=
\Pi_{\mathrm{seg}}(y_{ij})
=
\frac{\mathbf{u}_{ij}}{\|\mathbf{u}_{ij}\|_1}$.
This projection maps a dense pixel-wise mask to a fixed-length vector, and the resulting $\mathbf{s}_{ij}$ captures the relative class composition of the predicted segmentation.

Taken together, these instantiations show that $\Pi_{\mathrm{task}}$ maps heterogeneous task outputs into a shared semantic-presence space by retaining only class-level presence information while discarding task-specific structural details. This enables unified inner-group consistency computation and decision aggregation, as defined in Sec.~\ref{sec:method}.

\section{Preliminaries}
\label{sec:prelim}
We consider a dataset
$\mathcal{D}=\{(x_n,y_n^{(t)})\}_{n=1}^{|\mathcal{D}|}$,
where $x_n\in\mathcal{X}$ is an input image, and
$y_n^{(t)}$ is the corresponding annotation for task $t$.
The annotation format is task-dependent, thus varying across tasks. In this paper, we focus on settings in which different tasks share a common underlying class-level semantic space of $C$ classes, indexed by $\{1, 2, \ldots,C\}$, while exposing heterogeneous output formats.
Below, we formalize the outputs for three representative vision tasks used in our experiments.

\textbf{\emph{Image classification.}}
An image classification model predicts semantic classes and returns a discrete label set:
\begin{equation}
f^{cls}:\mathcal{X}\rightarrow \mathcal{P}(\{1,2,\ldots,C\}),
\end{equation}
where $\mathcal{P}(\{1,2,\ldots,C\})$ denotes the set of all subsets of $\{1,2,\ldots,C\}$. Each element in $f^{cls}(x)$ is the predicted probability that an object belonging to the corresponding class is present in image $x$.

\textbf{\emph{Object detection.}}
An object detection model returns a variable-length set of instance predictions:
\begin{equation}
f^{det}:\mathcal{X}\rightarrow \{(b_n,c_n)\}_{n=1}^{N_x},
\end{equation}
where $b_n\in\mathbb{R}^4$ is the predicted bounding box (e.g., in $(x_1,y_1,x_2,y_2)$ or $(x,y,w,h)$ format),
$c_n\in\{1,2,\ldots,C\}$ is the predicted class,
and $N_x$ is the number of returned detections for image $x$.

\textbf{\emph{Semantic segmentation.}}
A segmentation model predicts dense per-pixel semantics and returns a discrete label mask over the spatial grid:
\begin{equation}
f^{seg}:\mathcal{X}\rightarrow \{1,2, \ldots,C\}^{H\times W},
\end{equation}
where $H$ and $W$ are the height and width of the output mask, respectively, and each pixel at location $(h,w)$ is assigned a class label in $\{1,2,\ldots,C\}$.

\section{Applicability Discussion}
\label{sec:applicability}

In this paper, we identify a new model modification attack---cross-task model repurposing---that has not been explored in prior work. Existing approaches assume that the suspect model operates on the same task as the protected model, and are thus not designed to handle models repurposed for different tasks. To this end, we develop UniReFP to generate and verify fingerprints even when the suspect model is a repurposed version of the protected model. More specifically, UniReFP generates fingerprints in a unified semantic space and verifies the ownership of a suspect model based on unified semantic-presence evidence. This design naturally raises the question of when such semantic evidence is available, which we discuss below.

UniReFP focuses on cross-task repurposing where the task changes but the underlying semantic space remains shared. In other words, the protected and suspect models may have different output formats, but they describe the same set of semantic categories. If the semantic space also changes, for example, when a protected model trained on animal categories is repurposed to a vehicle-related task, the semantic-presence evidence used by UniReFP is no longer aligned between the protected and suspect models. Therefore, UniReFP is not expected to be effective in this setting. To the best of our knowledge, no existing model fingerprinting methods can address this challenging scenario as well. We identify this challenge and leave it for future work.

\section{Related Works}
\subsection{Model Ownership Protection}
Current model ownership protection techniques are broadly categorized into two families: watermarking-based~\cite{TYWS18,PIPD18,RWNNE19,PAND20} and fingerprinting-based methods \citep{IPIP21,DNNF21,FDNN22,UWSD24,geng2026irisfp,BUNNF25}.
The former safeguards model ownership by embedding an ownership signal through a training process and eliciting a predefined behavior corresponding to the signal~\cite{CNNWR22,MNNW23, RMWD24,RMWIP24,TRNNF25,Chainmark25}, while the latter protects ownership by exploiting unique characteristics of a model to craft fingerprints and verifying fingerprints to determine ownership. However, embedding an ownership signal into a model may degrade model utility and introduce additional security risks ~\cite{EPDA19,CWBD23}. Therefore, considering that model fingerprinting establishes ownership without modifying the model, it has been consider a promising technique for model ownership protection.

\subsection{Model Fingerprinting}
Most fingerprinting methods are designed for image classification tasks~\cite{IPIP21,DNNF21,IERF21,FDNN22,MVAM22,TMFA22,MFDN22,MCGF24,UWSD24,QURD25,BUNNF25}. Some approaches construct fingerprints from benign images and use the model’s response patterns (e.g., hard-sample behaviors or output statistics) as ownership evidence. While simple, such methods may offer limited separability when different models behave similarly on natural inputs. For example, Maho et al.~\cite{FCBI23} select a small set of benign images that maximize inter-model disagreement, using the resulting responses as a behavioral signature. Guan~\cite{AYSM22} computes an output correlation matrix on selected hard queries and verifies ownership by comparing this matrix with that of the protected model. Some approaches construct fingerprints using adversarial examples that induce model-specific behaviors. Cao et al.~\cite{IPIP21} generate adversarial inputs near decision boundaries as distinctive fingerprints. Extending this idea, Xu et al.~\cite{UWSD24} construct adversarial trajectories—sequences of perturbed inputs—that traverse boundary regions and capture richer model-specific behaviors. Godinot et al.~\cite{QURD25} show that initializing adversarial fingerprints from naturally misclassified inputs is effective. Chen et al.~\cite{TMFA22} perform teacher-classifier attribution by crafting fingerprint pairs and identifying the model whose knowledge best preserves pairwise response consistency. Zhang et al.~\cite{BUNNF25} improve fingerprint uniqueness under limited query budgets by selecting informative triggers that maximize mutual information, while Yang et al.~\cite{yang26anafp} optimize fingerprint queries using analytically guided objectives to enhance robustness and uniqueness.

While fingerprinting for image classification has been extensively studied, extending it to tasks such as object detection and semantic segmentation remains largely underexplored. Pan et al.~\cite{MVAM22} propose a task-agnostic framework that jointly learns fingerprint queries and a meta-verifier, enabling support for non-classification models by requiring only matched input/output dimensions. Yang et al.~\cite{yang26liteguard} further improve task-agnostic fingerprinting under limited reference models by leveraging checkpoint-based diversity and local verifiers for each fingerprint. However, these methods still assume that the suspect model preserves the same output structure as the protected model. When a pirated model is repurposed to a different task (e.g., from classification to detection or segmentation), the output space changes, rendering such verification mechanisms ineffective.

\section{The Computing Platform}
\label{sec:appendix_computingplatform}
We conduct all experiments on a high-performance server with an NVIDIA A100 GPU (40\,GB memory) and an Intel Xeon Gold 6246 CPU at 3.30\,GHz. Our implementation uses Python 3.9.21, PyTorch 2.5.1, and CUDA 12.1.

\section{Societal Impacts}
\label{sec:appendix_societalimpacts}

We propose UniReFP, a unified fingerprinting framework for verifying the ownership of deep neural network models under cross-task model repurposing. By enabling black-box ownership verification even when a stolen model is repurposed for a different task, our method contributes positively to intellectual property protection and trustworthy AI deployment. It can help discourage unauthorized model reuse and commercial misuse, thereby supporting fair attribution in both academia and industry.


\newpage
\section*{NeurIPS Paper Checklist}

\begin{enumerate}

\item {\bf Claims}
    \item[] Question: Do the main claims made in the abstract and introduction accurately reflect the paper's contributions and scope?
    \item[] Answer: \answerYes{} 
    \item[] Justification: The abstract and introduction accurately state the paper's main contribution, which is presented in Section~\ref{sec:method} and evaluated in Section~\ref{sec:exp}.
    \item[] Guidelines:
    \begin{itemize}
        \item The answer \answerNA{} means that the abstract and introduction do not include the claims made in the paper.
        \item The abstract and/or introduction should clearly state the claims made, including the contributions made in the paper and important assumptions and limitations. A \answerNo{} or \answerNA{} answer to this question will not be perceived well by the reviewers. 
        \item The claims made should match theoretical and experimental results, and reflect how much the results can be expected to generalize to other settings. 
        \item It is fine to include aspirational goals as motivation as long as it is clear that these goals are not attained by the paper. 
    \end{itemize}

\item {\bf Limitations}
    \item[] Question: Does the paper discuss the limitations of the work performed by the authors?
    \item[] Answer: \answerYes{} 
    \item[] Justification: The paper discusses the applicability and limitations in Appendix~\ref{sec:applicability}.
    \item[] Guidelines:
    \begin{itemize}
        \item The answer \answerNA{} means that the paper has no limitation while the answer \answerNo{} means that the paper has limitations, but those are not discussed in the paper. 
        \item The authors are encouraged to create a separate ``Limitations'' section in their paper.
        \item The paper should point out any strong assumptions and how robust the results are to violations of these assumptions (e.g., independence assumptions, noiseless settings, model well-specification, asymptotic approximations only holding locally). The authors should reflect on how these assumptions might be violated in practice and what the implications would be.
        \item The authors should reflect on the scope of the claims made, e.g., if the approach was only tested on a few datasets or with a few runs. In general, empirical results often depend on implicit assumptions, which should be articulated.
        \item The authors should reflect on the factors that influence the performance of the approach. For example, a facial recognition algorithm may perform poorly when image resolution is low or images are taken in low lighting. Or a speech-to-text system might not be used reliably to provide closed captions for online lectures because it fails to handle technical jargon.
        \item The authors should discuss the computational efficiency of the proposed algorithms and how they scale with dataset size.
        \item If applicable, the authors should discuss possible limitations of their approach to address problems of privacy and fairness.
        \item While the authors might fear that complete honesty about limitations might be used by reviewers as grounds for rejection, a worse outcome might be that reviewers discover limitations that aren't acknowledged in the paper. The authors should use their best judgment and recognize that individual actions in favor of transparency play an important role in developing norms that preserve the integrity of the community. Reviewers will be specifically instructed to not penalize honesty concerning limitations.
    \end{itemize}

\item {\bf Theory assumptions and proofs}
    \item[] Question: For each theoretical result, does the paper provide the full set of assumptions and a complete (and correct) proof?
    \item[] Answer: \answerNA{} 
    \item[] Justification: The paper does not include theoretical results. 
    \item[] Guidelines:
    \begin{itemize}
        \item The answer \answerNA{} means that the paper does not include theoretical results. 
        \item All the theorems, formulas, and proofs in the paper should be numbered and cross-referenced.
        \item All assumptions should be clearly stated or referenced in the statement of any theorems.
        \item The proofs can either appear in the main paper or the supplemental material, but if they appear in the supplemental material, the authors are encouraged to provide a short proof sketch to provide intuition. 
        \item Inversely, any informal proof provided in the core of the paper should be complemented by formal proofs provided in appendix or supplemental material.
        \item Theorems and Lemmas that the proof relies upon should be properly referenced. 
    \end{itemize}

    \item {\bf Experimental result reproducibility}
    \item[] Question: Does the paper fully disclose all the information needed to reproduce the main experimental results of the paper to the extent that it affects the main claims and/or conclusions of the paper (regardless of whether the code and data are provided or not)?
    \item[] Answer: \answerYes{} 
    \item[] Justification: We provide all the information needed to reproduce the main experimental results in Section~\ref{sec:method}, \ref{sec:exp}, and Appendix~\ref{sec:appendix_implementation_details}.
    \item[] Guidelines:
    \begin{itemize}
        \item The answer \answerNA{} means that the paper does not include experiments.
        \item If the paper includes experiments, a \answerNo{} answer to this question will not be perceived well by the reviewers: Making the paper reproducible is important, regardless of whether the code and data are provided or not.
        \item If the contribution is a dataset and\slash or model, the authors should describe the steps taken to make their results reproducible or verifiable. 
        \item Depending on the contribution, reproducibility can be accomplished in various ways. For example, if the contribution is a novel architecture, describing the architecture fully might suffice, or if the contribution is a specific model and empirical evaluation, it may be necessary to either make it possible for others to replicate the model with the same dataset, or provide access to the model. In general. releasing code and data is often one good way to accomplish this, but reproducibility can also be provided via detailed instructions for how to replicate the results, access to a hosted model (e.g., in the case of a large language model), releasing of a model checkpoint, or other means that are appropriate to the research performed.
        \item While NeurIPS does not require releasing code, the conference does require all submissions to provide some reasonable avenue for reproducibility, which may depend on the nature of the contribution. For example
        \begin{enumerate}
            \item If the contribution is primarily a new algorithm, the paper should make it clear how to reproduce that algorithm.
            \item If the contribution is primarily a new model architecture, the paper should describe the architecture clearly and fully.
            \item If the contribution is a new model (e.g., a large language model), then there should either be a way to access this model for reproducing the results or a way to reproduce the model (e.g., with an open-source dataset or instructions for how to construct the dataset).
            \item We recognize that reproducibility may be tricky in some cases, in which case authors are welcome to describe the particular way they provide for reproducibility. In the case of closed-source models, it may be that access to the model is limited in some way (e.g., to registered users), but it should be possible for other researchers to have some path to reproducing or verifying the results.
        \end{enumerate}
    \end{itemize}

\item {\bf Open access to data and code}
    \item[] Question: Does the paper provide open access to the data and code, with sufficient instructions to faithfully reproduce the main experimental results, as described in supplemental material?
    \item[] Answer: \answerYes{} 
    \item[] Justification: We provide code with sufficient instructions to faithfully reproduce the experimental results in the Supplemental Material.
    \item[] Guidelines:
    \begin{itemize}
        \item The answer \answerNA{} means that paper does not include experiments requiring code.
        \item Please see the NeurIPS code and data submission guidelines (\url{https://neurips.cc/public/guides/CodeSubmissionPolicy}) for more details.
        \item While we encourage the release of code and data, we understand that this might not be possible, so \answerNo{} is an acceptable answer. Papers cannot be rejected simply for not including code, unless this is central to the contribution (e.g., for a new open-source benchmark).
        \item The instructions should contain the exact command and environment needed to run to reproduce the results. See the NeurIPS code and data submission guidelines (\url{https://neurips.cc/public/guides/CodeSubmissionPolicy}) for more details.
        \item The authors should provide instructions on data access and preparation, including how to access the raw data, preprocessed data, intermediate data, and generated data, etc.
        \item The authors should provide scripts to reproduce all experimental results for the new proposed method and baselines. If only a subset of experiments are reproducible, they should state which ones are omitted from the script and why.
        \item At submission time, to preserve anonymity, the authors should release anonymized versions (if applicable).
        \item Providing as much information as possible in supplemental material (appended to the paper) is recommended, but including URLs to data and code is permitted.
    \end{itemize}

\item {\bf Experimental setting/details}
    \item[] Question: Does the paper specify all the training and test details (e.g., data splits, hyperparameters, how they were chosen, type of optimizer) necessary to understand the results?
    \item[] Answer: \answerYes{} 
    \item[] Justification: The experimental details are provided in Section~\ref{sec:exp} and the Appendix~\ref{sec:appendix_implementation_details}.
    \item[] Guidelines:
    \begin{itemize}
        \item The answer \answerNA{} means that the paper does not include experiments.
        \item The experimental setting should be presented in the core of the paper to a level of detail that is necessary to appreciate the results and make sense of them.
        \item The full details can be provided either with the code, in appendix, or as supplemental material.
    \end{itemize}

\item {\bf Experiment statistical significance}
    \item[] Question: Does the paper report error bars suitably and correctly defined or other appropriate information about the statistical significance of the experiments?
    \item[] Answer: \answerYes{} 
    \item[] Justification: The paper reports results with mean and standard deviation over five runs. 
    \item[] Guidelines:
    \begin{itemize}
        \item The answer \answerNA{} means that the paper does not include experiments.
        \item The authors should answer \answerYes{} if the results are accompanied by error bars, confidence intervals, or statistical significance tests, at least for the experiments that support the main claims of the paper.
        \item The factors of variability that the error bars are capturing should be clearly stated (for example, train/test split, initialization, random drawing of some parameter, or overall run with given experimental conditions).
        \item The method for calculating the error bars should be explained (closed form formula, call to a library function, bootstrap, etc.)
        \item The assumptions made should be given (e.g., Normally distributed errors).
        \item It should be clear whether the error bar is the standard deviation or the standard error of the mean.
        \item It is OK to report 1-sigma error bars, but one should state it. The authors should preferably report a 2-sigma error bar than state that they have a 96\% CI, if the hypothesis of Normality of errors is not verified.
        \item For asymmetric distributions, the authors should be careful not to show in tables or figures symmetric error bars that would yield results that are out of range (e.g., negative error rates).
        \item If error bars are reported in tables or plots, the authors should explain in the text how they were calculated and reference the corresponding figures or tables in the text.
    \end{itemize}

\item {\bf Experiments compute resources}
    \item[] Question: For each experiment, does the paper provide sufficient information on the computer resources (type of compute workers, memory, time of execution) needed to reproduce the experiments?
    \item[] Answer: \answerYes{} 
    \item[] Justification: We report information about our computing platform in the Appendix~\ref{sec:appendix_computingplatform}.
    \item[] Guidelines:
    \begin{itemize}
        \item The answer \answerNA{} means that the paper does not include experiments.
        \item The paper should indicate the type of compute workers CPU or GPU, internal cluster, or cloud provider, including relevant memory and storage.
        \item The paper should provide the amount of compute required for each of the individual experimental runs as well as estimate the total compute. 
        \item The paper should disclose whether the full research project required more compute than the experiments reported in the paper (e.g., preliminary or failed experiments that didn't make it into the paper). 
    \end{itemize}
    
\item {\bf Code of ethics}
    \item[] Question: Does the research conducted in the paper conform, in every respect, with the NeurIPS Code of Ethics \url{https://neurips.cc/public/EthicsGuidelines}?
    \item[] Answer: \answerYes{} 
    \item[] Justification: The research conforms to the NeurIPS Code of Ethics.
    \item[] Guidelines:
    \begin{itemize}
        \item The answer \answerNA{} means that the authors have not reviewed the NeurIPS Code of Ethics.
        \item If the authors answer \answerNo, they should explain the special circumstances that require a deviation from the Code of Ethics.
        \item The authors should make sure to preserve anonymity (e.g., if there is a special consideration due to laws or regulations in their jurisdiction).
    \end{itemize}

\item {\bf Broader impacts}
    \item[] Question: Does the paper discuss both potential positive societal impacts and negative societal impacts of the work performed?
    \item[] Answer: \answerYes{} 
    \item[] Justification:  We provide a discussion on the societal impacts of our work in the Appendix~\ref{sec:appendix_societalimpacts}.
    \item[] Guidelines:
    \begin{itemize}
        \item The answer \answerNA{} means that there is no societal impact of the work performed.
        \item If the authors answer \answerNA{} or \answerNo, they should explain why their work has no societal impact or why the paper does not address societal impact.
        \item Examples of negative societal impacts include potential malicious or unintended uses (e.g., disinformation, generating fake profiles, surveillance), fairness considerations (e.g., deployment of technologies that could make decisions that unfairly impact specific groups), privacy considerations, and security considerations.
        \item The conference expects that many papers will be foundational research and not tied to particular applications, let alone deployments. However, if there is a direct path to any negative applications, the authors should point it out. For example, it is legitimate to point out that an improvement in the quality of generative models could be used to generate Deepfakes for disinformation. On the other hand, it is not needed to point out that a generic algorithm for optimizing neural networks could enable people to train models that generate Deepfakes faster.
        \item The authors should consider possible harms that could arise when the technology is being used as intended and functioning correctly, harms that could arise when the technology is being used as intended but gives incorrect results, and harms following from (intentional or unintentional) misuse of the technology.
        \item If there are negative societal impacts, the authors could also discuss possible mitigation strategies (e.g., gated release of models, providing defenses in addition to attacks, mechanisms for monitoring misuse, mechanisms to monitor how a system learns from feedback over time, improving the efficiency and accessibility of ML).
    \end{itemize}
    
\item {\bf Safeguards}
    \item[] Question: Does the paper describe safeguards that have been put in place for responsible release of data or models that have a high risk for misuse (e.g., pre-trained language models, image generators, or scraped datasets)?
    \item[] Answer: \answerNA{} 
    \item[] Justification: Our work does not involve the release of models or datasets that present high risk for misuse.
    \item[] Guidelines:
    \begin{itemize}
        \item The answer \answerNA{} means that the paper poses no such risks.
        \item Released models that have a high risk for misuse or dual-use should be released with necessary safeguards to allow for controlled use of the model, for example by requiring that users adhere to usage guidelines or restrictions to access the model or implementing safety filters. 
        \item Datasets that have been scraped from the Internet could pose safety risks. The authors should describe how they avoided releasing unsafe images.
        \item We recognize that providing effective safeguards is challenging, and many papers do not require this, but we encourage authors to take this into account and make a best faith effort.
    \end{itemize}

\item {\bf Licenses for existing assets}
    \item[] Question: Are the creators or original owners of assets (e.g., code, data, models), used in the paper, properly credited and are the license and terms of use explicitly mentioned and properly respected?
    \item[] Answer: \answerYes{} 
    \item[] Justification: All assets used in this work are publicly available and properly cited, and we use them in accordance with their respective licenses and terms of use.
    \item[] Guidelines:
    \begin{itemize}
        \item The answer \answerNA{} means that the paper does not use existing assets.
        \item The authors should cite the original paper that produced the code package or dataset.
        \item The authors should state which version of the asset is used and, if possible, include a URL.
        \item The name of the license (e.g., CC-BY 4.0) should be included for each asset.
        \item For scraped data from a particular source (e.g., website), the copyright and terms of service of that source should be provided.
        \item If assets are released, the license, copyright information, and terms of use in the package should be provided. For popular datasets, \url{paperswithcode.com/datasets} has curated licenses for some datasets. Their licensing guide can help determine the license of a dataset.
        \item For existing datasets that are re-packaged, both the original license and the license of the derived asset (if it has changed) should be provided.
        \item If this information is not available online, the authors are encouraged to reach out to the asset's creators.
    \end{itemize}

\item {\bf New assets}
    \item[] Question: Are new assets introduced in the paper well documented and is the documentation provided alongside the assets?
    \item[] Answer: \answerYes{} 
    \item[] Justification: We release our implementation in the supplemental material, including the code for fingerprint generation, ownership verification, and evaluation.
    \item[] Guidelines:
    \begin{itemize}
        \item The answer \answerNA{} means that the paper does not release new assets.
        \item Researchers should communicate the details of the dataset\slash code\slash model as part of their submissions via structured templates. This includes details about training, license, limitations, etc. 
        \item The paper should discuss whether and how consent was obtained from people whose asset is used.
        \item At submission time, remember to anonymize your assets (if applicable). You can either create an anonymized URL or include an anonymized zip file.
    \end{itemize}

\item {\bf Crowdsourcing and research with human subjects}
    \item[] Question: For crowdsourcing experiments and research with human subjects, does the paper include the full text of instructions given to participants and screenshots, if applicable, as well as details about compensation (if any)? 
    \item[] Answer: \answerNA{} 
    \item[] Justification: The paper does not involve crowdsourcing experiments or research with human subjects.
    \item[] Guidelines:
    \begin{itemize}
        \item The answer \answerNA{} means that the paper does not involve crowdsourcing nor research with human subjects.
        \item Including this information in the supplemental material is fine, but if the main contribution of the paper involves human subjects, then as much detail as possible should be included in the main paper. 
        \item According to the NeurIPS Code of Ethics, workers involved in data collection, curation, or other labor should be paid at least the minimum wage in the country of the data collector. 
    \end{itemize}

\item {\bf Institutional review board (IRB) approvals or equivalent for research with human subjects}
    \item[] Question: Does the paper describe potential risks incurred by study participants, whether such risks were disclosed to the subjects, and whether Institutional Review Board (IRB) approvals (or an equivalent approval/review based on the requirements of your country or institution) were obtained?
    \item[] Answer: \answerNA{} 
    \item[] Justification: The paper does not involve human-subject research or crowdsourcing. Therefore, IRB approval or equivalent review is not applicable.
    \item[] Guidelines:
    \begin{itemize}
        \item The answer \answerNA{} means that the paper does not involve crowdsourcing nor research with human subjects.
        \item Depending on the country in which research is conducted, IRB approval (or equivalent) may be required for any human subjects research. If you obtained IRB approval, you should clearly state this in the paper. 
        \item We recognize that the procedures for this may vary significantly between institutions and locations, and we expect authors to adhere to the NeurIPS Code of Ethics and the guidelines for their institution. 
        \item For initial submissions, do not include any information that would break anonymity (if applicable), such as the institution conducting the review.
    \end{itemize}

\item {\bf Declaration of LLM usage}
    \item[] Question: Does the paper describe the usage of LLMs if it is an important, original, or non-standard component of the core methods in this research? Note that if the LLM is used only for writing, editing, or formatting purposes and does \emph{not} impact the core methodology, scientific rigor, or originality of the research, declaration is not required.
    \item[] Answer: \answerNA{} 
    \item[] Justification: The core method, experiments, and scientific claims do not involve LLMs.
    \item[] Guidelines:
    \begin{itemize}
        \item The answer \answerNA{} means that the core method development in this research does not involve LLMs as any important, original, or non-standard components.
        \item Please refer to our LLM policy in the NeurIPS handbook for what should or should not be described.
    \end{itemize}

\end{enumerate}

\end{document}